\documentclass{IEEEtran}
\usepackage{cite}
\usepackage{amsmath,amssymb,amsfonts}
\usepackage{algorithmic}
\usepackage{graphicx}
\usepackage{textcomp}
\usepackage{multirow}
\usepackage{hyperref}
\usepackage{multicol}
\usepackage{url}
\usepackage{booktabs}
\usepackage{makecell}
\usepackage{color}
\usepackage{bbding}

\def\BibTeX{{\rm B\kern-.05em{\sc i\kern-.025em b}\kern-.08em
		T\kern-.1667em\lower.7ex\hbox{E}\kern-.125emX}}
\ifCLASSINFOpdf
\else
\fi

\begin{document}
	\title{MvKSR: Multi-view Knowledge-guided Scene Recovery for Hazy and Rainy Degradation}
	\author{Dong Yang, Wenyu Xu, Yuan Gao, Yuxu Lu, Jingming Zhang, and Yu Guo
	\thanks{The work described in this paper is supported by a grant from the Research Grants Council of the Hong Kong Special Administrative Region, China (Grant Number. PolyU 15201722). D. Yang and W. Xu are co-first authors. Corresponding author: Yuxu Lu)}
    \thanks{D. Yang and Y. Lu are with the Department of Logistics and Maritime Studies, the Hong Kong Polytechnic University, Hong Kong. (e-mail: dong.yang@polyu.edu.hk, yuxulouis.lu@connect.polyu.hk).}
    \thanks{Y. Gao is with the School of Navigation, Wuhan University of Technology, Wuhan 430063, China (e-mail: yuangao@whut.edu.cn).}	
    \thanks{W. Xu is with the Wuhan Baosight Software Co., Ltd, Wuhan 430063, China (e-mail: xuwenyu\_971652@baosight.com).}
    \thanks{J. Zhang is with the Intelligent Transportation Systems Research Center, Wuhan University of Technology, Wuhan 430063, China(e-mail: jeremy.zhang@whut.edu.cn).}
    \thanks{Y. Guo is with the School of Computing and Information Systems, Singapore Management University, Singapore (e-mail: yuguo65896@gmail.com).}

}
	
\maketitle
	
\begin{abstract}
    High-quality imaging is crucial for ensuring safety supervision and intelligent deployment in fields like transportation and industry. It enables precise and detailed monitoring of operations, facilitating timely detection of potential hazards and efficient management. However, adverse weather conditions, such as atmospheric haziness and precipitation, can have a significant impact on image quality. When the atmosphere contains dense haze or water droplets, the incident light scatters, leading to degraded captured images. This degradation is evident in the form of image blur and reduced contrast, increasing the likelihood of incorrect assessments and interpretations by intelligent imaging systems (IIS). To address the challenge of restoring degraded images in hazy and rainy conditions, this paper proposes a novel multi-view knowledge-guided scene recovery network (termed MvKSR). Specifically, guided filtering is performed on the degraded image to separate high/low-frequency components. Subsequently, an en-decoder-based multi-view feature coarse extraction module (MCE) is used to coarsely extract features from different views of the degraded image. The multi-view feature fine fusion module (MFF) will learn and infer the restoration of degraded images through mixed supervision under different views. Additionally, we suggest an atrous residual block to handle global restoration and local repair in hazy/rainy/mixed scenes. Extensive experimental results demonstrate that MvKSR outperforms other state-of-the-art methods in terms of efficiency and stability for restoring degraded scenarios in IIS. The source code is available at \url{https://github.com/LouisYuxuLu/MvKSR}.

\end{abstract}


\section{Introduction}
    \IEEEPARstart{I}{ntelligent} imaging systems (IIS) provide crucial value in transportation \cite{zhou2022real} and industry \cite{yang2020visual} by offering real-time monitoring, early warning, and data analysis capabilities, contributing to the development and operation of society. However, during periods of atmospheric haziness and rainfall, the presence of water droplets or water vapor particles causes the scattering of incoming light, leading to diminished visibility and a decrease in the distinction between objects of interest and the surrounding environment in images \cite{cao2021two}. The decline in image quality poses obstacles to the detection, recognition, and tracking functionalities of IIS \cite{gao2023let}. Therefore, it is crucial for imaging equipment to possess the capability to adapt to adverse weather conditions and exhibit enhanced perceptual abilities in order to ensure the security and effectiveness of IIS. At present, image dehazing and deraining methods can be classified into two main categories: traditional and learning-based methods. Traditional physical model-based methods demonstrate satisfactory performance in dealing with uncomplicated scenarios. Learning-based methods excel in achieving enhanced accuracy, robustness, and generalizability in image restoration tasks by acquiring knowledge about mapping relationships from extensive datasets.
    Traditional dehazing methods, such as dark channel priors (DCP) \cite{he2010single} and haze density estimation \cite{yeh2013haze}, primarily employ atmospheric scattering model (ASM) to estimate the latent clear feature in degraded images. These methods \cite{kim2019fast,liu2022rank} assume that the scattering in the image is a result of haze, aiming to restore image clarity by estimating and removing the haze. Traditional deraining methods mainly focus on image restoration by employing rain streaks models \cite{kang2011automatic,chen2013generalized,li2016rain}, which incorporate the motion and shape characteristics of rain streaks. By modeling rain streaks and estimating their trajectories and shapes, they mitigate the visual distortions caused by rainfall in the image. However, in practical scenarios, rainfall processes often coexist with haze \cite{cao2021two}. In such cases, it requires a joint establishment of degradation models that integrate atmospheric scattering and rain streaks effects to simulate the degradation process with better accuracy. However, the unpredictable degradation scenes make traditional methods often have unnatural visual restoration performance.
    \begin{figure*}[t]
        \centering
        \setlength{\abovecaptionskip}{0.cm}
        \includegraphics[width=1.00\linewidth]{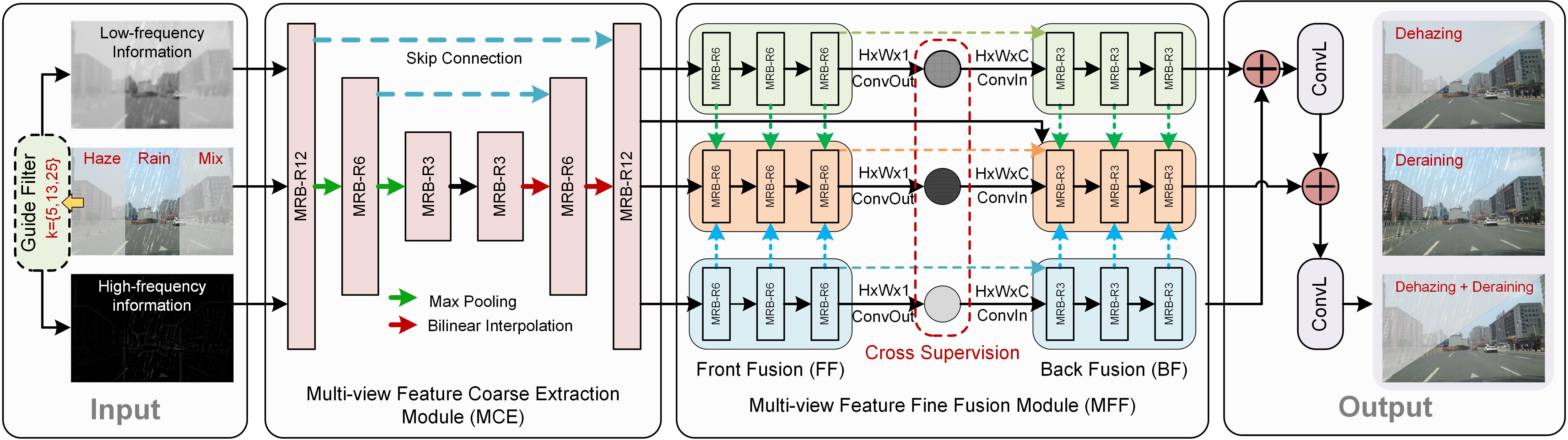}
            \caption{The flowchart of the multi-view knowledge-guided scene recovery network (termed MvKSR). Multi-view feature coarse extraction module (MCE) will perform en-decoder based learning inference on the degraded image and the corresponding high/low-frequency components. Multi-view feature fine fusion module (MFF) (including front and back fusion) will guide the restoration of degraded images through mixed supervision.}
        \label{Figure_Flowchart}
    \end{figure*}
    Various classic learning-based methods have been developed for image dehazing and deraining tasks. The end-to-end methods, which use convolutional neural networks (CNNs) \cite{luo2023rain, guo2023scanet, qu2023deep,yin2023multiscale} or Transformer \cite{song2023vision,liang2022drt} to directly learn the mapping relationship between input images and ground truth. Generative adversarial networks (GANs) \cite{dong2020fd,zhang2019image} simultaneously train a generator and a discriminator to restore the degraded image. The generator converts the input image into a clean image, while the discriminator evaluates the realism of the generated image. This adversarial training mechanism enhances the quality and visual effect of the generated images. Multi-scale-based methods \cite{jiang2020multi,li2021multi} can capture richer information about haze and rain streaks. These methods often utilize the feature pyramid structure or attention mechanism of CNNs or Transformer to restore the details and structure of images effectively. Additionally, physical model-based methods combine physical models with deep learning to more accurately infer and eliminate the effects of haze and rain lines \cite{shen2018deep,wang2020deep,hu2021single}. By incorporating the atmospheric scattering model and rainfall properties into the image recovery process, these methods provide more precise results.
    The coexistence of haze and rain can be effectively described by the physical imaging model, which captures the degradation process. Nevertheless, the inequitable dispersion of haze and sporadic precipitation patterns give rise to disparate levels of degradation in the image's depth of field, intricate elements, and color rendition. Therefore, the physical imaging models may not be fully appropriate in these particular cases. This work proposes a novel multi-view knowledge-guided scene recovery network (termed MvKSR), which aims to enhance the correlation between hazy and rainy imaging in complicated degraded scene restoration. MvKSR can effectively allocate acquired information from intricate degradation scenarios to particular tasks. Firstly, the degraded image is subjected to guided filtering \cite{he2012guided} to separate its high/low-frequency components. Subsequently, the multi-view feature coarse extraction module (MCE) will perform en-decoder-based learning and inference on the coarse features of the three views (i.e., degraded image and the corresponding high/low-frequency components). To guide the restoration process, the multi-view feature fine fusion module (MFF) (including front and back fusion) will guide the restoration of degraded images through mixed supervision under different views in the grayscale domain. In addition, an atrous residual module is suggested to handle global restoration and local repair in hazy/rainy/mixed scenes. The main contributions of this work can be summarized as follows 
	\begin{itemize}
		\item We propose a novel multi-view knowledge-guided scene recovery network (MvKSR) to achieve hazy, rainy, and mixed scene recovery in IIS. Specifically, MvKSR can learn different imaging degradations from multiple views through MCE and MFF to obtain positive performance gains in complex degraded scene restoration.
		\item We propose a mixed supervision strategy, which greatly improves the generalization ability of the network by calculating the loss of multi-view output in the grayscale domain through full supervision and self-supervision.
		\item Both quantitative and qualitative experiments demonstrate that our MvKSR has the capacity to improve the visual quality of images significantly in different weather conditions. In addition, it could prove to be efficient with lower computational cost than state-of-the-art methods, which has significant industrial application value in IIS.
	\end{itemize}
    The remainder of this work is organized as follows. Degradation models for hazy, rainy, and mixed images are given in Section \ref{sec:pim}. Our MvKSR is detailedly described in Section \ref{sec:mvksr}. Experimental results and discussion are provided in Section \ref{sec:exper}. Section \ref{sec:conc} finally presents the conclusion.

\section{Physical Imaging Models}\label{sec:pim}
    The atmosphere scattering model (ASM) is a widely used approximation for simulating the haze effect in images. Mathematically, it can be expressed as
    \begin{equation}\label{pim:haze}
    	I^{h}(x)=I^{c}(x) t(x)+A(1-t(x)),
    \end{equation}
    where $I^{h}(x)$ and $I^{c}(x)$ are the intensity of pixel $x$ of the hazy image $I^{h}$ and potential clear image $I^{c}$, respectively. $t(x)$ is the transmission map, and $A$ is the global atmospheric light. The transmission map $t(x) = e^{-\beta d(x)}$, where $\beta$ is the atmosphere scattering coefficient and $d(x)$ represents the scene depth.
    To model the rain effect, a rain-streak layer $S(x)$ can be superimposed on the clear image $I^{c}$, which can be given as
    \begin{equation}\label{pim:rain}
    	I^{r}(x)=I^{c}(x)+S(x),
    \end{equation}
    where $I^{r}(x)$ is the intensity of pixel $x$ of rainy image $I^{r}$.
    In heavy rain conditions, the visual degradation of distant scenes can be similar to the haze effect, which is known as the rain veiling effect. To simulate this effect, the ASM can be integrated with the rain-streak-based model, i.e.,
    \begin{equation}\label{pim:mix}
    	I^{m}(x)=(I^{c}(x) + S(x)) t(x)+ A(1-t(x)),
    \end{equation}
    where $I^{m}(x)$ is the intensity of pixel $x$ of rain-haze image $I^{m}$. Accurate physical imaging models are crucial for knowledge-driven methods to better understand image degradation processes. In this work, we use existing paired clean images and degraded images synthesized based on Eq. \ref{pim:haze}, \ref{pim:rain}, and \ref{pim:mix} as training datasets to jointly optimize the scene restoration performance of the network.

\section{Proposed Method}\label{sec:mvksr}
\subsection{Guided Filter Decomposition}
    \begin{figure}[t]
        \centering
        \setlength{\abovecaptionskip}{0.cm}
        \includegraphics[width=1.00\linewidth]{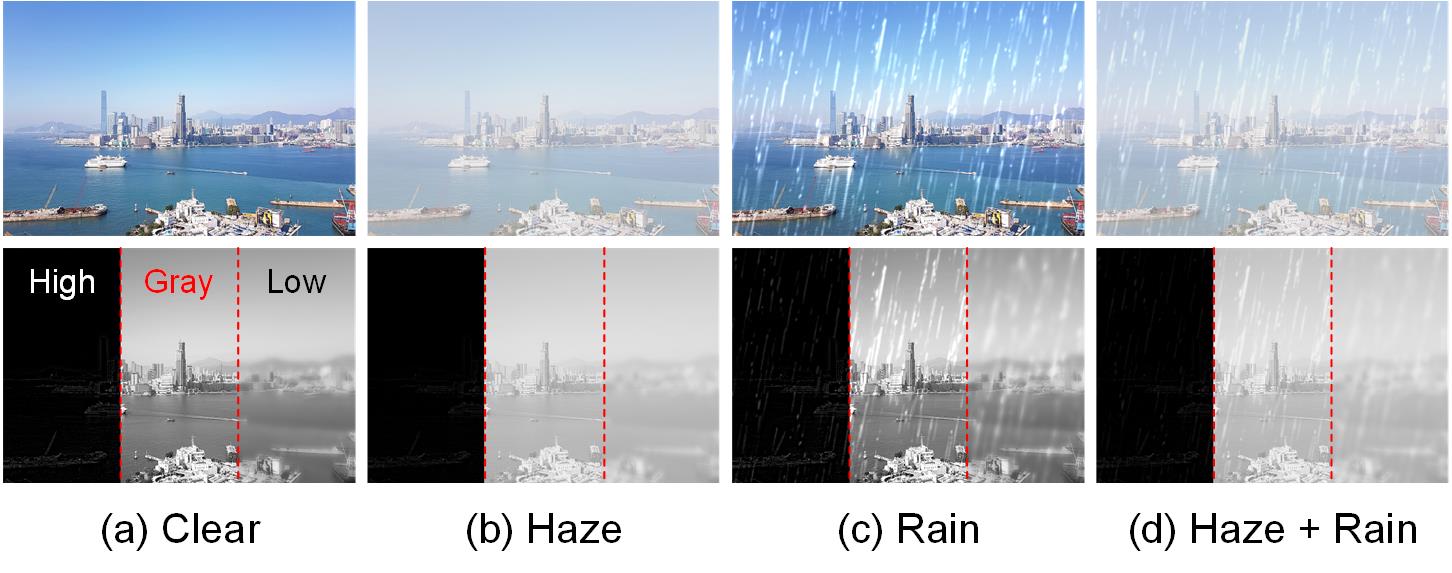}
        \caption{Clear and synthetic different types of degraded images and corresponding high- and low-frequency components.}
        \label{Figure03}
    \end{figure}
    The high-frequency layer of an image contains scene textures, while the low-frequency layer contains scene layout/structure. The low-frequency layer is affected by haze due to its smoothness, while the high-frequency layer is disrupted by rain (see Fig. \ref{Figure03}). We separate the high and low frequencies because they represent different aspects of the degraded scene, and thus, they require different handling. Therefore, we first convert the RGB color image $I_c$ to grayscale domain, i.e.,
    \begin{equation}
        I_g = 0.229 \cdot I_c^r + 0.587 \cdot I_c^g + 0.114 \cdot I_c^b,   
    \end{equation}
    where $I_g$ is the grayscale image. $I_c^r$, $I_c^g$, and $I_c^b$ are the components of  $I_c$ in the three channels. We extract and repair the high- and low-frequency information of degraded images in the grayscale domain for two reasons. Firstly, irregular and random rain streaks can severely degrade the original image during rainy conditions and introduce unwanted high-frequency artifacts. Since rain streaks usually have lighter shades, the grayscale domain enables better separation of the original image from the undesired high-frequency features. Secondly, in real-world hazy images, the strength of low-frequency information differs across different color channels. However, potential low-frequency features can be extracted and mapped relatively easily in the grayscale domain.
    In this work, we suggest the guided filtering algorithm \cite{he2012guided} to decompose the degraded grayscale image $I_g$ into low-frequency layer $I_l$ and the corresponding high-frequency layer $I_h$, which can be given as 
    \begin{equation}
        \left\{\begin{array}{l}
            I_l= G_{k,\epsilon}(I_g, I_p) \\
            I_h= 1- I_l
        \end{array}\right.,
    \end{equation}
    where $G_{k,\epsilon}(\cdot)$ represents the guided filtering operation, $k$ and $\epsilon=0.1$ are the parameters that decide the filter kernel size and blur degree of the guided filtering, respectively. It should be noted that we use the degraded image itself as the guide image (i.e., $I_g=I_p$). To reduce potential information loss caused by inappropriate selection of input image parameters, we suggest to generate three high-frequency layers and three corresponding low-frequency layers with different filter kernel size (i.e., $k=\{5,13,25\}$).

\subsection{Atrous Residual Block}
    \begin{figure}[t]
        \centering
        \setlength{\abovecaptionskip}{0.cm}
        \includegraphics[width=1.00\linewidth]{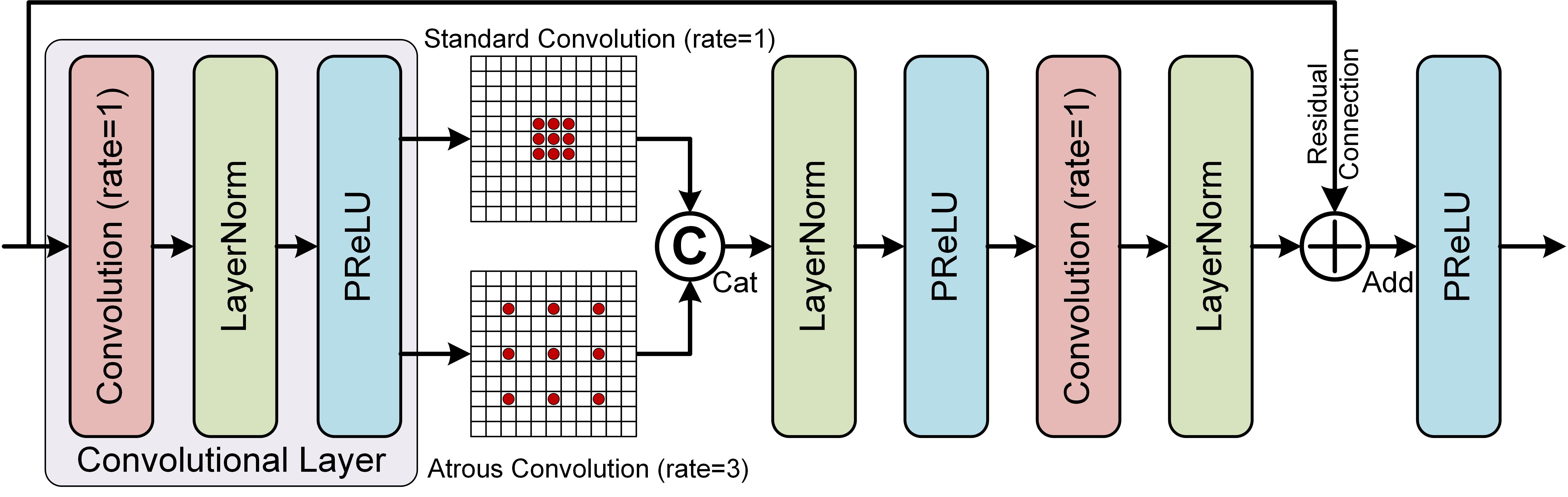}
        \caption{The pipeline of proposed standard and atrous mixed residual block (MRB) with atrous rate $r=3$.}
        \label{Figure04}
    \end{figure}
    Residual learning \cite{he2016deep} has demonstrated its efficient performance in different fields of computer vision. To increase the receptive field of the residual block to better extract valuable features from degraded scenes, as shown in Fig. \ref{Figure04}, we propose a standard and atrous mixed residual block (MRB, $\textbf{MRB}(\cdot)$) as the basic learning unit of the proposed network and consists of mixed convolutional layers ($\textbf{MCL}(\cdot)$), which consists of two types of convolution operation i.e., standard convolution $\mathcal{C}(\cdot)$ and atrous convolution  $\mathcal{C}^r(\cdot)$ with atrous rate $r$, i.e.,
    \begin{equation}
        \textbf{MCL}(x_{mcl})=\mathcal{P}(\mathcal{N}(cat[\mathcal{C}^r(x_{mcl});\mathcal{C}(x_{mcl})])),
    \end{equation}
    where $x_{mcl}$ is the input of learning block, $\mathcal{N}(\cdot)$ and $\mathcal{P}(\cdot)$ represent layer normalization (LayerNorm) and parametric rectified linear unit (PReLu). The concatenation of standard and atrous convolution can achieve a balance between spatial detail information extraction and global perception. Therefore, MRB can be defined as 
    \begin{equation}
        \textbf{MRB}(x_{mcl})=\mathcal{P}(\mathcal{N}(\mathcal{C}(\textbf{MCL}(\textbf{MCL}(x_{mcl}))+x_{mcl}).
    \end{equation}
    As the basic learning module, MRB will be suggested to construct a scene recovery network with stable enhancement performance and low computational cost. In addition, this work suggests three different types of MRB with varying atrous rates, i.e., $r=12, 6, 3$. The feature map is downsampled and the network propagates deeper, the underlying features are gradually refined. Decreasing atrous rates can mitigate the loss of texture details by reducing the void rate.

\subsection{Multi-view Feature Coarse Extraction Module}
    Degraded RGB color images and corresponding multi-layer high/low-frequency information will be concatenated as the input of the multi-view feature coarse extraction module (MCE). As shown in Fig. \ref{Figure_Flowchart}, the en-decoder network, as the main framework for the MCE, can map to a low-dimensional space through a series of convolution and pooling operations in the encoding stage to capture advanced features from multiple viewpoints. By introducing degraded multi-view information into the decoding process through skip connections, the correct structure and color information can be ensured. To prevent the weakening of texture details caused by too high hole rates when features are downsampled, we set the atrous rates of the MRBs in the en-decoder at $12$, $6$, and $3$, with channel numbers of $16$, $32$, and $64$, respectively. MCE will provide rich high-frequency, low-frequency, grayscale-domain, and color-domain features for the subsequent module, enabling the restoration of low-quality images with complex degradation factors through mixed supervision in multiple viewpoints. 

\subsection{Multi-view Feature Fine Fusion Module}\label{ss:mff}
\begin{figure}[t]
    	\centering
    	\setlength{\abovecaptionskip}{0.cm}
    	\includegraphics[width=1.00\linewidth]{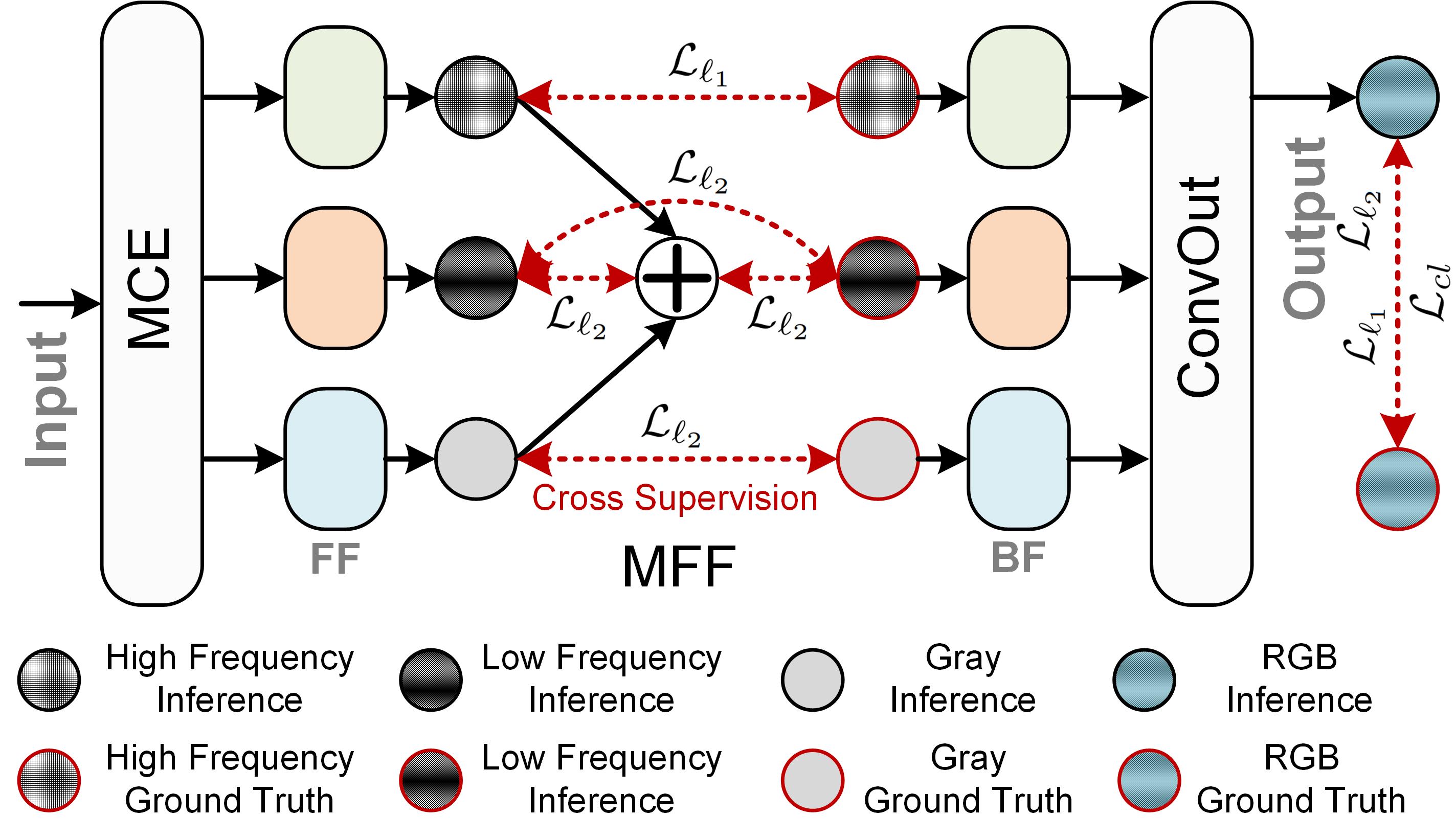}
    	\caption{Cross supervision in multi-view feature fine fusion module (MFF).}
    	\label{Figure_MFF}
    \end{figure}

\begin{figure}[t]
        \centering
        \setlength{\abovecaptionskip}{0.cm}
        \includegraphics[width=1.00\linewidth]{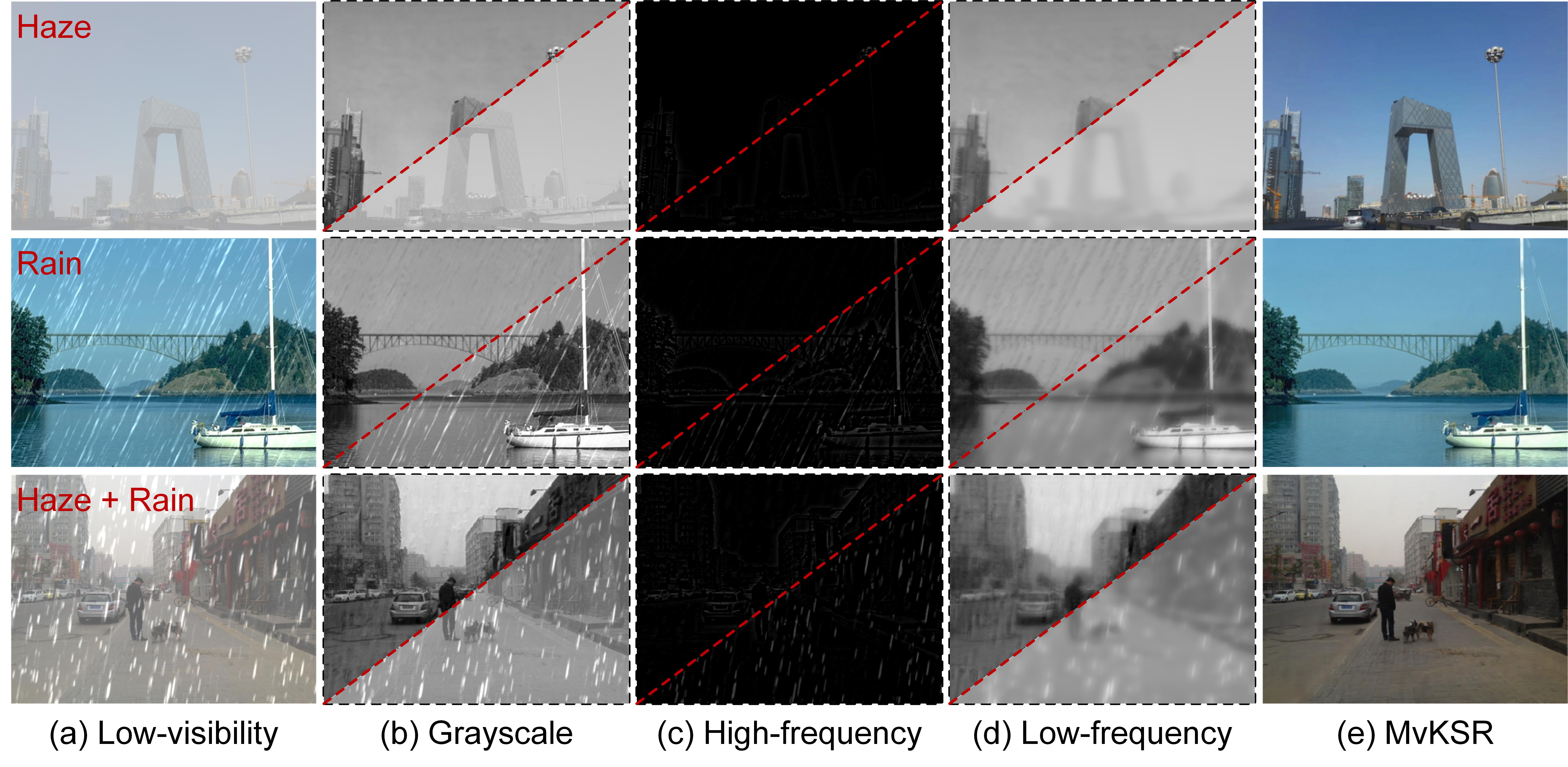}
        \caption{Example of the scene recovery of MFF and MvKSR under different views. The lower triangles in (b)-(d) are degraded patterns, and the corresponding restored patterns by our method are shown in the upper triangles.}
        \label{Figure05}
    \end{figure}

    The deep model can gain a more comprehensive understanding of the intrinsic structure and relationships within the data by analyzing data from different views. As shown in Fig. \ref{Figure_MFF}, the proposed multi-view feature fine fusion module (MFF) mainly consists of three parts, i.e., front fusion (FF), cross supervision, and back fusion (BF). In FF, we will finely decompose the output of the last layer of the CEM decoder and generate grayscale domain feature $I_g^f$, high-frequency feature $I_h^f$, and low-frequency feature $I_l^f$. Cross supervision will perform full supervision of $I_g^f$, $I_h^f$, and $I_l^f$ with their corresponding ground truth $I_g^c$, $I_h^c$, and $I_l^c$ and self-supervision of $I_g^f$ and ($I_h^f + I_l^f$). Therefore, we perform joint optimization learning through $\ell_1$-norm and $\ell_2$-norm, that is, the cross-supervised loss function $\mathcal{L}_{cs}$ can be given as
    \begin{equation}
            \begin{aligned}
                \mathcal{L}_{cs} = \|I_g^f-I_g^c\|_2 +  \|I_h^f-I_h^c\|_1 & + \|I_l^f-I_l^c\|_2\\ &+ \|I_g^f-(I_h^f + I_l^f)\|_2.
        \end{aligned}
    \end{equation}
    The MFF-generated multi-view features have equal importance, so we give the same weight value to the sub-loss functions in $\mathcal{L}_{cs}$. BF will further fuse the multi-view features generated by FF and MCE to further optimize feature parameter learning. As shown in Fig. \ref{Figure05}, in the process of network learning or reasoning, high- and low-frequency features are beneficial to enhance grayscale domain features.
\subsection{Ultimate Loss Function}
    To supervise the learning process of the network, we choose multi-scale structural similarity (MS-SSIM) loss $\mathcal{L}_{\text{MS-SSIM}}$ and contrastive regularization (CR) loss $\mathcal{L}_{cr}$ as training objectives to drive the model optimization. Therefore, the total loss $\mathcal{L}_{\text {total}}$ of proposed MvKSR can be defined as 
    \begin{equation}
    	\mathcal{L}_{total}=\lambda_1\mathcal{L}_{\text{MS-SSIM}} + \lambda_2\mathcal{L}_{cr},
    \end{equation}
    where $\lambda_1$ and $\lambda_2$ denote the weight value for each loss term, respectively. Extensive experimental results show that $\lambda_1=0.8$ and $\lambda_2=0.2$ will have the best quantitative and qualitative performance. The fluctuation of the weight in a small range will not produce obvious changes in the image restoration results. However, when the weight of CR is too large, the performance of MvKSR will become worse.
    \subsubsection{Multi-Scale Structural Similarity Loss}
    The MS-SSIM loss can help to guide the model to generate or enhance images that are visually similar to the target images, improving the perceptual quality of the output, which can be given as
    \begin{equation}
        \mathcal{L}_{\text{MS-SSIM}} = 1 - \prod_{i=1}^{N} \left( \text{{SSIM}}(I_r, I_{gt}) \right)^{\alpha_i},
    \end{equation}
    where $I_r$ and $I_{gt}$ are the restored image and corresponding ground truth. $\text{{SSIM}}(I_r, I_{gt})$ represents the structural similarity (SSIM) loss between $I_r$ and $I_{gt}$. $N$ is the number of scales in the multi-scale decomposition. $\alpha_i$ is the weight assigned to the $i$-th scale. The SSIM can be defined as
    \begin{equation}
        \text{SSIM} = \frac{{(2\mu_{I_r}\mu_{I_{gt}} + c_1)(2\sigma{{I_r}{I_{gt}}} + c_2)}}{{(\mu_{I_r}^2 + \mu_{I_{gt}}^2 + c_1)(\sigma_{I_r}^2 + \sigma_{I_{gt}}^2 + c_2)}},
    \end{equation}
    where $\mu_{I_r}$ and $\mu_{I_{gt}}$ are the average values of ${I_r}$ and ${I_{gt}}$ respectively. $\sigma_{I_r}$ and $\sigma_{I_{gt}}$ are the standard deviations of ${I_r}$ and ${I_{gt}}$ respectively. $\sigma_{{I_r}{I_{gt}}}$ is the covariance between ${I_r}$ and ${I_{gt}}$. $c_1$ and $c_2$ are constants to stabilize the division. The MS-SSIM loss function calculates the SSIM between the input images at multiple scales and combines them using the weights $\alpha_i$. This allows the loss function to capture both local and global structural similarities between the images.
    \subsubsection{Contrastive Regularization Loss}
    The CR loss is beneficial for MvKSR to learn from positive pairs and separate representations between negative pairs. It can guide the network to focus on the relevant features and discard irrelevant or noisy information. By incorporating CR into the training process, the network can learn a more discriminative and robust representation of the data. Therefore, the CR loss $\mathcal{L}_{cr}$ can be given as
    \begin{equation}
    	\mathcal{L}_{cr}= \sum_{i=1}^n \omega_i \cdot \frac{\left\|\psi_i\left(I_r\right)-\psi_i(I_{gt})\right\|_1}{\left\|\psi_i(I)-\psi_i(I_{gt})\right\|_1},
    \end{equation}
    where $\psi_i(\cdot),. i=1,2, \cdots n$, refer to extracting the $i$-th hidden features from the VGG-19 network pre-trained. $\omega_i$ are weight coefficients, and we set $\omega_1=\frac{1}{32}$, $\omega_2=\frac{1}{16}$, $\omega_3=\frac{1}{8}$, $\omega_4=\frac{1}{4}$, and $\omega_5=1$. More implementation details about the CR loss can be found in literature \cite{wu2021contrastive}.
    \setlength{\tabcolsep}{2.00pt}
    \begin{table}[tb]
        \centering
        \caption{PSNR, SSIM, FSIM, and VSI results of various methods on hazy scene recovery. The best results are in \textbf{bold}, and the second best are with \underline{underline}.}
        \begin{tabular}{l|cccc}
        \hline
        Methods          & PSNR $\uparrow$ & SSIM $\uparrow$ & FSIM $\uparrow$ & VSI $\uparrow$ \\ \hline\hline
        DCP \cite{he2010single}              & 16.49$\pm$3.62 & 0.801$\pm$0.083 & 0.942$\pm$0.027 & 0.977$\pm$0.011 \\ 
        MSCNN \cite{ren2016single}            & 19.09$\pm$5.48 & 0.856$\pm$0.125 & 0.932$\pm$0.068 & 0.977$\pm$0.022 \\ 
        FFANet \cite{qin2020ffa}           & 18.27$\pm$5.19 & 0.829$\pm$0.130 & 0.903$\pm$0.089 & 0.973$\pm$0.027 \\ 
        ROP+ \cite{liu2022rank}              & 18.62$\pm$3.30 & 0.855$\pm$0.064 & 0.935$\pm$0.034 & 0.973$\pm$0.013 \\ 
        TOENet \cite{gao2023let}           & \underline{22.70$\pm$3.96} & \underline{0.922$\pm$0.070} & \underline{0.970$\pm$0.026} & \underline{0.986$\pm$0.011} \\ 
        MIRNet \cite{zamir2020learning}           & 18.86$\pm$5.84 & 0.826$\pm$0.136 & 0.913$\pm$0.077 & 0.972$\pm$0.029 \\ 
        AirNet \cite{li2022all}           & 18.34$\pm$4.78 & 0.797$\pm$0.134 & 0.896$\pm$0.079 & 0.964$\pm$0.031 \\ 
        TransWeather \cite{valanarasu2022transweather}     & 22.06$\pm$6.26 & 0.890$\pm$0.093 & 0.950$\pm$0.052 & 0.984$\pm$0.018 \\ 
        WeatherDiff \cite{ozdenizci2023restoring}      & 17.14$\pm$3.69 & 0.831$\pm$0.101 & 0.924$\pm$0.047 & 0.971$\pm$0.021 \\ 
        WGWSNet \cite{zhu2023learning}          & 19.20$\pm$5.12 & 0.892$\pm$0.100 & 0.957$\pm$0.049 & 0.983$\pm$0.018 \\ \hline
        MvKSR            & \textbf{27.10$\pm$6.56} &  \textbf{0.950$\pm$0.055} &  \textbf{0.986$\pm$0.018} &  \textbf{0.995$\pm$0.007} \\ \hline
        \end{tabular}\label{Table_metrics_haze}
    \end{table}    
    \setlength{\tabcolsep}{2.00pt}
    \begin{table}[tb]
    	\centering
    	\caption{PSNR, SSIM, FSIM, and VSI results of various methods on rainy scene recovery. The best results are in \textbf{bold}, and the second best are with \underline{underline}.}
    	        \begin{tabular}{l|cccc}
        \hline
        Methods          & PSNR $\uparrow$ & SSIM $\uparrow$ & FSIM $\uparrow$ & VSI $\uparrow$ \\ \hline\hline
        DDN \cite{fu2017removing}         & 27.95$\pm$1.85 & 0.818$\pm$0.056 & 0.900$\pm$0.038 & 0.975$\pm$0.011 \\ 
        DID \cite{zhang2018density}          & 23.70$\pm$1.60 & 0.803$\pm$0.052 & 0.897$\pm$0.030 & 0.973$\pm$0.009 \\ 
        LPNet \cite{fu2019lightweight}        & 31.63$\pm$3.24 & 0.892$\pm$0.041 & 0.937$\pm$0.026 & 0.986$\pm$0.008 \\ 
        DiG \cite{ran2020single}          & 31.28$\pm$3.06 & 0.892$\pm$0.034 & 0.949$\pm$0.017 & 0.989$\pm$0.005 \\ 
        DualGCN \cite{fu2021rain}      & \textbf{36.04$\pm$4.88} & \underline{0.965$\pm$0.028} & \underline{0.979$\pm$0.017} & \underline{0.995$\pm$0.005} \\ 
        MIRNet \cite{zamir2020learning}       & 23.67$\pm$3.33 & 0.735$\pm$0.092 & 0.872$\pm$0.042 & 0.965$\pm$0.015 \\ 
        AirNet \cite{li2022all}       & 20.60$\pm$2.84 & 0.694$\pm$0.091 & 0.861$\pm$0.042 & 0.962$\pm$0.015 \\ 
        TransWeather \cite{valanarasu2022transweather} & 25.08$\pm$2.82 & 0.822$\pm$0.071 & 0.910$\pm$0.035 & 0.978$\pm$0.013 \\ 
        WeatherDiff \cite{ozdenizci2023restoring}  & 20.71$\pm$2.28 & 0.804$\pm$0.087 & 0.895$\pm$0.044 & 0.967$\pm$0.019 \\ 
        WGWSNet \cite{zhu2023learning}      & 28.64$\pm$2.13 & 0.836$\pm$0.066 & 0.906$\pm$0.040 & 0.976$\pm$0.013 \\ \hline
        MvKSR        & \underline{35.85$\pm$2.58} & \textbf{0.975$\pm$0.010} & \textbf{0.986$\pm$0.007} & \textbf{0.997$\pm$0.002} \\ \hline
        \end{tabular}\label{Table_metrics_rain}
    \end{table}    
    \setlength{\tabcolsep}{2.00pt}
    \begin{table}[!ht]
    	\centering
    	\caption{PSNR, SSIM, FSIM, and VSI results of various methods on mixed scene (Hazy+Rainy) recovery. The best results are in \textbf{bold}, and the second best are with \underline{underline}.}
            \begin{tabular}{l|cccc}
            \hline
            Methods          & PSNR $\uparrow$ & SSIM $\uparrow$ & FSIM $\uparrow$ & VSI $\uparrow$ \\ \hline\hline
            MIRNet \cite{zamir2020learning}       & 18.42$\pm$4.64 & 0.715$\pm$0.095 & 0.858$\pm$0.054 & 0.959$\pm$0.022 \\ 
            AirNet \cite{li2022all}       & 19.26$\pm$4.18 & 0.673$\pm$0.093 & 0.842$\pm$0.061 & 0.951$\pm$0.025 \\ 
            TransWeather \cite{valanarasu2022transweather} & \underline{21.87$\pm$4.92} & \underline{0.788$\pm$0.083} & \underline{0.894$\pm$0.048} & \underline{0.971$\pm$0.018} \\ 
            WeatherDiff \cite{ozdenizci2023restoring}  & 17.11$\pm$3.17 & 0.745$\pm$0.087 & 0.872$\pm$0.045 & 0.956$\pm$0.019 \\ 
            WGWSNet \cite{zhu2023learning}      & 17.68$\pm$4.38 & 0.748$\pm$0.087 & 0.872$\pm$0.054 & 0.962$\pm$0.019 \\ \hline
            MvKSR        & \textbf{28.02$\pm$5.39} & \textbf{0.928$\pm$0.058} & \textbf{0.972$\pm$0.020} & \textbf{0.992$\pm$0.007} \\ \hline
            \end{tabular}\label{Table_metrics_mix}
    \end{table}    

\section{Experiments and Discussions}\label{sec:exper}
    This section provides an overview of the network implementation, including the evaluation metrics, dataset, and configuration of network parameters. We showcase the benefits of MvKSR for image restoration and computational time with other competive methods. We also validate the significance of the component modules through ablation experiments.
    %

    %
    %

    %
    \begin{figure*}[htb]
        \centering
        \includegraphics[width=1.00\linewidth]{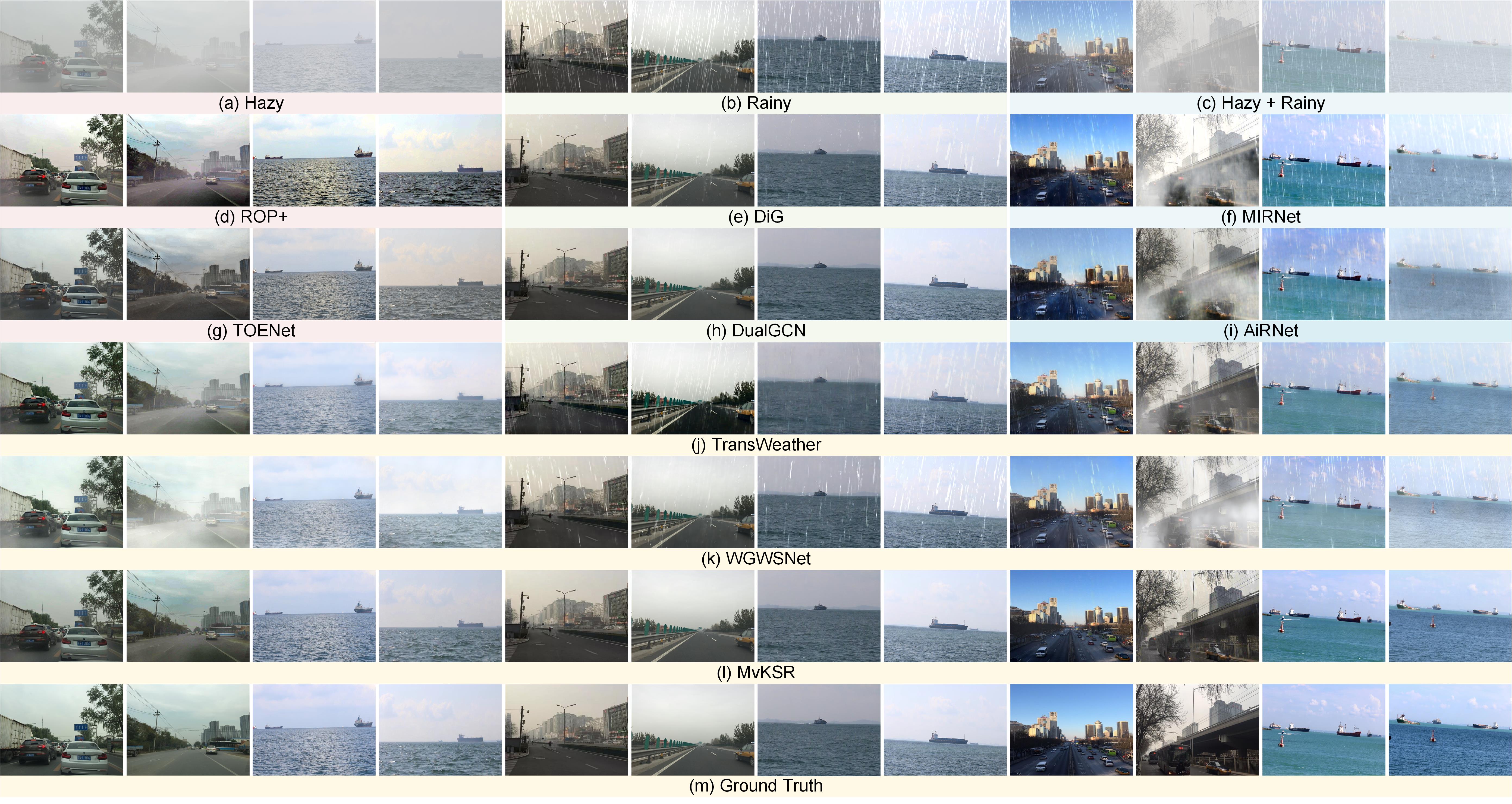}
        \caption{Visual comparisons of hazy/rainy/mixed (hazy+rainy) scene recovery from RESIDE-OTS \cite{li2018benchmarking} and SMD \cite{prasad2017video}. (a) Hazy, (b) Rainy, (c) Hazy + Rainy, restored images, generated by (d) ROP+ \cite{liu2022rank}, (e) DiG \cite{ran2020single}, (f) MIRNet \cite{zamir2020learning}, (g) TOENet \cite{gao2023let}, (h) DualGCN \cite{fu2021rain}, (i) AirNet \cite{li2022all}, (j) TransWeather \cite{valanarasu2022transweather}, (k) WGWSNet \cite{zhu2023learning}, (l) MvKSR, and (m) Ground Truth, respectively.}
        \label{Fig_haze_rain_mix}
    \end{figure*}
    \begin{figure}[htb]
        \centering
        \includegraphics[width=1.00\linewidth]{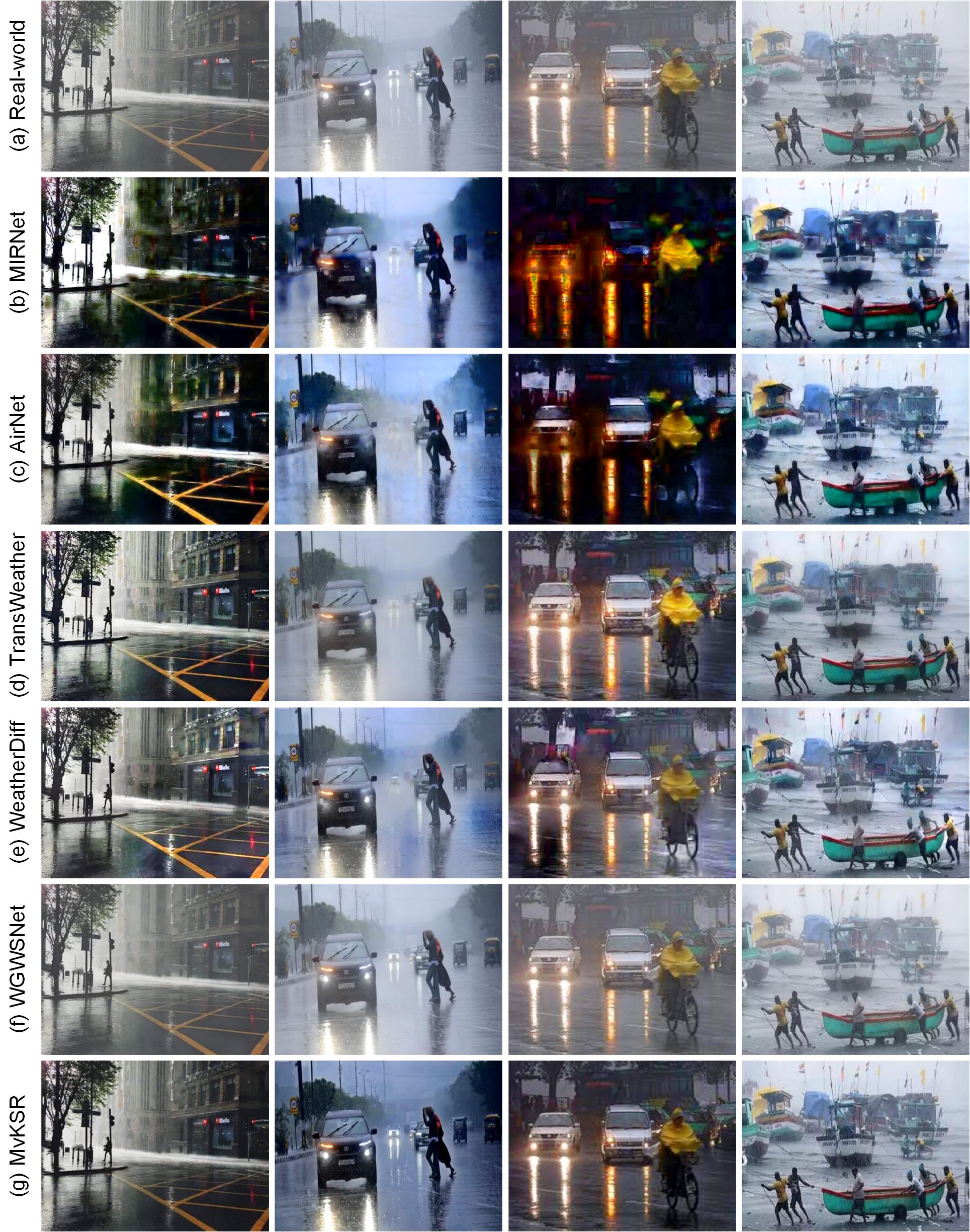}
        \caption{Visual comparisons of scene recovery performance from real-world low-visibility images. (a) Real-world low-visibility images, restored images, generated by (b) MIRNet \cite{zamir2020learning}, (c) AirNet \cite{li2022all}, (d) TransWeather \cite{valanarasu2022transweather}, (e) WeatherDiff \cite{ozdenizci2023restoring}, (f) WGWSNet \cite{zhu2023learning}, and (g) MvKSR, respectively.}
        \label{Fig_real}
    \end{figure}

\subsection{Dataset and Implementation Details}
    The focal points of the application scenarios discussed in this work pertain to the imaging of terrestrial and oceanic environments. Consequently, the training dataset comprises RESIDED-OTS (which incorporates depth information) \cite{li2018benchmarking} for land scenes, the Singapore Maritime Dataset (SMD) \cite{prasad2017video} for water scenes. We selected 2000 images each from RESIDED-OTS and SMD as the training dataset and an additional 500 images each for performance analysis of MvKSR and other competitive methods. The atmospheric light value of the damaged image is artificially extracted in real-world conditions. These values are then used in conjunction with an atmospheric scattering model (i.e., Eq. \ref{pim:haze}) to generate more realistic hazy images. The synthesis of degraded images that closely resemble real-world conditions can address the limitation of insufficient paired training datasets, hence enhancing the restoration performance of the network. And the rain steaks released by \cite{yang2019joint} are then applied with the Eq. \ref{pim:rain} to generate rainy images. In the meantime, the mixed (hazy+rainy) degraded images are generated by Eq. \ref{pim:mix}. The initial learning rate is set to 0.001 and is reduced by one-tenth after every 30 epochs. The training and testing of the MvKSR model are performed in a Python 3.7 environment using the PyTorch software package. The hardware setup includes an Intel(R) Core(TM) i9-13900K CPU running at 5.80GHz and an Nvidia GeForce RTX 3080 GPU.

\subsection{Competitive Methods and Evaluation Metrics}
    To evaluate the effectiveness of the proposed method, we conduct a comparative analysis between MvKSR and various state-of-the-art methods, which encompass both traditional and learning-based. For the dehazing task, we select DCP \cite{he2010single}, MSCNN \cite{ren2016single}, FFANet \cite{qin2020ffa}, ROP+ \cite{liu2022rank}, and TOENet \cite{gao2023let} as the compared methods. For the deraining task, we select DDN \cite{fu2017removing}, DID \cite{zhang2018density}, LPNet \cite{fu2019lightweight}, DIG \cite{ran2020single}, and DualGCN \cite{fu2021rain}. For the mixed scene (haze+rain) recovery, we select MIRNet \cite{zamir2020learning}, AirNet \cite{li2022all}, TransWeather \cite{valanarasu2022transweather}, WeatherDiff \cite{ozdenizci2023restoring}, and WGWSNet \cite{zhu2023learning}. Furthermore, to ensure the integrity and objectivity of the experiments, all compared methods are obtained exclusively from the author's source code.
    To ensure a precise evaluation of the enhancement and recovery capabilities of various methods, we selected a comprehensive set of evaluation metrics, which includes peak signal-to-noise ratio (PSNR), structure similarity index measure (SSIM) \cite{wang2004image}, feature similarity index measure (FSIM) \cite{zhang2011fsim}, and visual saliency-induced index (VSI) \cite{zhang2014vsi}. A higher value for each of the four metrics indicates a superior performance in the context of image recovery.

\subsection{Synthetic Quantitative and Visual Analysis}
    We conducted a visual analysis of degraded images with references to assess the generalization ability of MvKSR in different scenarios. We select classic low-visibility images of land and ocean under various degradation scenarios.

\subsubsection{Dehazing}
    We first compute objective evaluation metrics for the test images from RESIDE-OTS \cite{li2018benchmarking} and SMD \cite{prasad2017video}. As shown in Table \ref{Table_metrics_haze}, MvKSR ranks first in all evaluation indicators. To assess the dehazing performance of MvKSR, we present a visual comparison in Fig. \ref{Fig_haze_rain_mix} (Columns one to four). Notably, RESIDE-OTS, which incorporates depth information, produces low-visibility images that closely resemble real degraded images. However, traditional methods encounter difficulties in obtaining depth information from the degraded image, leading to either excessive enhancement of local areas with a loss of texture details or insufficient enhancement, allowing degradation factors to persist. We also conducted unbiased tests on SMD, where depth information was not considered. Most methods struggled to extract valuable information from these degraded images accurately. The complexity of the imaging environment often deviates from the original degradation imaging model, making it challenging for traditional methods to achieve satisfactory visual restoration performance across different degradation scenes. Furthermore, learning-based methods heavily rely on training data and lack generalizability, making it difficult to achieve satisfactory visual restoration results in unpredictable imaging degradation environments. In our evaluations, DCP exhibited excessively high saturation, MSCNN and FFANet failed to completely remove haze. ROP+ and TOENet are yellowish, which may be caused by their methods considering sandstorm image recovery. In general, MvKSR combines the advantages of both network models and physical prior information, enabling it to achieve satisfactory visual performance in dehazing task.

\subsubsection{Deraining}
    Similar to the dehazing experiment, we first calculate objective evaluation metrics for RESIDE-OTS \cite{li2018benchmarking} and SMD \cite{prasad2017video}, as presented in Table \ref{Table_metrics_rain}. In the case of RESIDE-OTS, MvKSR achieves the top rank in all metrics. Regarding SMD, although MvKSR may not rank first in every indicator, it is important to note that MvKSR is a multi-scene enhancement method and still remains highly competitive compared to specialized rain removal methods. The visual effects of different deraining methods are shown in Fig. \ref{Fig_haze_rain_mix} (columns five to eight). DDN and DID demonstrate limited effectiveness in removing rain marks. On the other hand, LPNet and DIG exhibit the capability to remove a significant portion of the rain marks, although they may struggle when faced with dense rain marks or complex background scenes, leaving some marks unremoved. In comparison, DualGAN achieves a better rain removal effect, yet it still leaves a few rain marks in localized areas. In contrast, our MvKSR excels in performing image restoration tasks involving rain marks of varying densities and complex backgrounds, resulting in the most visually satisfying restoration effect.
\subsubsection{Mixed Scene (Hazey+Rainy) Recovery}
    Similar to the dehazing and deraining experiments, we initially compute objective evaluation metrics for the test images sourced from RESIDE-OTS \cite{li2018benchmarking} and SMD \cite{prasad2017video}. As illustrated in Table \ref{Table_metrics_mix}, MvKSR achieves the first place rankings across all evaluation criteria. Recovering mixed scenes poses a challenging task. Observing Fig. \ref{Fig_haze_rain_mix} (columns nine to twelve), it is evident that existing methods for mixed scene image recovery fail to deliver satisfactory results when both haze and rain are present simultaneously. This may be due to their methods not being specifically designed for IIS and lacking consideration of the distinctive characteristics of degradation in road and water traffic images. Additionally, poor generalization ability could contribute to their limitations. In contrast, our MvKSR is tailored for IIS scenarios and can still achieve remarkable results in mixed scene recovery within IIS.
\subsection{Real-world Visual Analysis}
    The real-world low-visibility imaging process is more complex in IIS. As shown in Fig \ref{Fig_real}, we selected four degraded images related to land/ocean for visual comparison. The images restored by MIRNet and AiRNet exhibit local color distortion. TransWeather performs well overall but cannot completely remove degradation effects when the haze density is too high. WeatherDiff restored images have local artifacts. WGWSNet cannot accurately extract valuable features from degraded images. In contrast, benefiting from multi-view degradation perception and joint optimization with cross supervision, MvKSR achieves the best visual performance.
\subsection{Ablation Analysis}\label{ss:aa}
\subsubsection{High/Low-Frequency Analysis}\label{ss:hlfaa}
    In this paper, we decompose images into high-frequency information and low-frequency information through guided filtering and assist in guiding the backbone network to recover potentially clear images from degraded scenes. As shown in Table \ref{table_hl}, through the guidance of high/low frequency prior information, the performance of the network is further improved. High/low frequency information can assist in separating unwanted rain streaks during the inferring process, and can further strengthen potential edge texture features, improving the restoration robustness of MvKSR in different degraded scenarios.
    \setlength{\tabcolsep}{7.0pt}
    \begin{table}[t]
	\centering
	\caption{Ablation analysis (PSNR / SSIM) of the suggested high and low frequency information.}
        \begin{tabular}{cc|ccc}
        \hline
        High & Low & Haze         & Rain         & Haze+Rain    \\ \hline\hline
             &     & 24.352 / 0.917 & 34.145 / 0.961 & 23.114 / 0.862 \\ 
        \CheckmarkBold    &     & 24.911 / 0.924 & 34.753 / 0.969 & 24.245 / 0.877 \\ 
             & \CheckmarkBold   & 25.034 / 0.926 & 35.108 / 0.973 & 24.623 / 0.881 \\ \hline
        \CheckmarkBold    & \CheckmarkBold   & 25.417 / 0.933 & 35.229 / 0.978 & 24.871 / 0.887 \\ \hline
        \end{tabular}\label{table_hl}
    \end{table}
    \setlength{\tabcolsep}{4.75pt}
    \begin{table}[t]
	\centering
	\caption{Ablation analysis (PSNR/SSIM) of the suggested loss function in MFF.}
        \begin{tabular}{ccc|ccc}
        \hline
         High & Low & Self         & Haze & Rain & Haze+Rain    \\ \hline\hline
             &  &  & 23.877 / 0.913 & 33.470 / 0.947& 22.145 / 0.845\\ 
         \CheckmarkBold    &  &  & 24.894 / 0.922 & 34.633 / 0.966& 23.775 / 0.866\\ 
         & \CheckmarkBold &  & 24.231 / 0.919 & 33.754 / 0.960& 22.884 /0.868\\ 
        \CheckmarkBold    & \CheckmarkBold &  & 25.108 / 0.931 & 34.514 / 0.974 & 24.476 / 0.884\\  \hline
         \CheckmarkBold    & \CheckmarkBold & \CheckmarkBold & 25.417 / 0.933 &  35.229 / 0.978 & 24.871 / 0.887\\  \hline
        \end{tabular}\label{table_aal}
    \end{table}
    \setlength{\tabcolsep}{0.80pt}
    \begin{table}[!ht]
    \centering
    \caption{Comparison of the model size and running time between MvKSR and other methods of the 720p image ($1080\times720$ pixels).}
    \begin{tabular}{c|l|cccc}
        \hline
        Task                                                                            & Methods      & Language   & Platform   & Size (KB) & Time (Sec.) \\ \hline \hline
        \multirow{5}{*}{Dehazing}                                                       & DCP \cite{he2010single}          & Matlab (C) & ---        & ---       & 1.164       \\
                                                                                        & MSCNN \cite{ren2016single}        & Matlab (C) & ---        & ---       & 0.838       \\
                                                                                        & FFANet \cite{qin2020ffa}       & Python (G) & Pytorch    & 25999     & 1.039       \\
                                                                                        & ROP+ \cite{liu2022rank}         & Matlab (C) & ---        & ---       & 0.138       \\
                                                                                        & TOENet \cite{gao2023let}       & Python (G) & Pytorch    & 2557      & 0.1       \\ \hline
        \multirow{5}{*}{Deraining}                                                      & DDN \cite{fu2017removing}          & Python (G) & Tensorflow & 228       & 0.343       \\
                                                                                        & DID \cite{zhang2018density}          & Python (G) & Pytorch    & 1513      & 0.029       \\
                                                                                        & LPNet \cite{fu2019lightweight}        & Python (G) & Tensorflow & 1513      & 0.232       \\
                                                                                        & DIG \cite{ran2020single}          & Matlab (C) & ---        & ---       & 2.922       \\
                                                                                        & DualGCN \cite{fu2021rain}      & Python (G) & Tensorflow & 10669     & 8.972       \\ \hline
        \multirow{7}{*}{\begin{tabular}[c]{@{}c@{}}Mixed\\ Scene\\ Recovery\end{tabular}} & MIRNet \cite{zamir2020learning}       & Python (G) & Pytorch    & 373510    & 0.612       \\
                                                                                        & AirNet \cite{li2022all}       & Python (G) & Pytorch    & 35393     & 0.642       \\
                                                                                        & TransWeather \cite{valanarasu2022transweather} & Python (G) & Pytorch    & 85669     & 0.007       \\
                                                                                        & WeatherDiff \cite{ozdenizci2023restoring}  & Python (G) & Pytorch    & 1296805   & 152.011     \\
                                                                                        & WGWSNet \cite{zhu2023learning}      & Python (G) & Pytorch    & 101921    & 0.982       \\  \cline{2-6}
                                                                              & MvKSR        & Python (G) & Pytorch    & 5681      & 0.077       \\ 
                                                                              & MvKSR-Fast    & Python (G) & Pytorch    & 5681      & 0.026       \\ \hline
    \end{tabular}
    \label{Table_time}
    \end{table}
\subsubsection{MFF Loss Analysis}\label{ss:mffla}
    Each component of the loss function in subsection \ref{ss:mff} fully exerts its important value in the learning phase of the network. As shown in Table \ref{table_aal}, the recovery performance of network is significantly improved by introducing supervision of high-frequency (High) and low-frequency (Low) information. Self-supervision (Self) can learn more diverse and useful feature representations, further enhancing the network's scene generalization ability.

\subsection{Running Time Comparisons}
    To demonstrate the computational efficiency of our MvKSR, we perform a comparative analysis of the running times of various methods. As indicated in Table \ref{Table_time}, the size of our MvKSR model is a mere 5681 KB. Moreover, it exhibits superior speed compared to most other methods, processing a 720P ($1080 \times 720$ pixels) image in just 0.077 seconds on a GPU. In addition, we still introduce fast guided filtering \cite{he2015fast} without compromising network performance, making it well-suited for deployment on edge devices in IIS applications.
    %

    %
    %

\section{Conclusion}\label{sec:conc}
    This paper proposes a multi-view knowledge-guided scene recovery network (termed MvKSR) for the intelligent imaging systems (IIS), which successfully restores hazy, rainy, and mixed degraded images by a single network. Specifically, we suggest to separating high-frequency and low-frequency components of degraded images by guided filtering. Subsequently, an en-decoder network is used to roughly extract features from different views of the degraded image. The multi-view feature fine fusion module will guide the restoration of degraded images through mixed supervision in the grayscale and RGB color domains. Additionally, an atrous residual block is employed to handle global restoration and local repair in hazy/rainy/mixed scenes. The comprehensive experimental results demonstrate that the proposed MvKSR outperforms other state-of-the-art methods in terms of efficiency and stability for image restoration task in IIS.

\ifCLASSOPTIONcaptionsoff
\newpage
\fi
\bibliographystyle{IEEEtran}
\bibliography{Ref.bib}
\tiny

\end{document}